\documentclass[10pt,journal,compsoc]{IEEEtran}

\ifCLASSOPTIONcompsoc
  \usepackage[nocompress]{cite}
\else
  \usepackage{cite}
\fi
\ifCLASSINFOpdf
\else
\fi
\usepackage{cite}
\usepackage{multirow}
\usepackage{graphicx}
\usepackage{subfigure}
\usepackage{color}
\usepackage{dcolumn}
\usepackage{amsmath}
\usepackage{multirow}
\usepackage{booktabs}
\usepackage[ruled,vlined]{algorithm2e}

\usepackage{verbatim}
\usepackage{float}
\usepackage{comment}

\begin{document}

\title{DiabDeep: Pervasive Diabetes Diagnosis based on Wearable Medical Sensors and Efficient Neural Networks}

\author{Hongxu Yin, Bilal Mukadam, Xiaoliang Dai and Niraj K. Jha,~\IEEEmembership{Fellow,~IEEE}
\thanks{This work was supported by NSF Grant No. CNS-1617640. Hongxu Yin, 
Bilal Mukadam, Xiaoliang Dai, and Niraj K. Jha are with the Department
of Electrical Engineering, Princeton University, Princeton,
NJ, 08544 USA, e-mail:\{hongxuy, bilal, xdai, jha\}@princeton.edu}}

\IEEEtitleabstractindextext{%
\begin{abstract}
Diabetes impacts the quality of life of millions of people around the globe. 
However, diabetes diagnosis is still an arduous process, given that 
this disease develops and gets treated outside the clinic. The emergence of 
wearable medical sensors (WMSs) and machine learning points to a potential way 
forward to address this challenge. WMSs enable a continuous, yet 
user-transparent, mechanism to collect and analyze physiological signals.
However, disease diagnosis based on WMS data and its effective deployment on 
resource-constrained edge devices remain challenging due to inefficient 
feature extraction and vast computation cost. To address these problems, we 
propose a framework called DiabDeep that combines efficient neural networks 
(called DiabNNs) with off-the-shelf WMSs for pervasive diabetes diagnosis. 
DiabDeep bypasses the feature extraction stage and acts directly on WMS data. 
It enables both an (i) accurate inference on the server, e.g., a desktop, and 
(ii) efficient inference on an edge device, e.g., a smartphone, to obtain a 
balance between accuracy and efficiency based on varying resource budgets and 
design goals. On the resource-rich server, we stack sparsely connected 
layers to deliver high accuracy. On the resource-scarce edge device, we use 
a hidden-layer long short-term memory based recurrent layer to 
substantially cut down on 
computation and storage costs while incurring only a minor accuracy loss. 
At the core of our system lies a grow-and-prune training flow: it leverages 
gradient-based growth and magnitude-based pruning algorithms to enable 
DiabNNs to learn both weights and connections, while improving accuracy and 
efficiency. We demonstrate the effectiveness of DiabDeep through a detailed 
analysis of data collected from 52 participants. For server (edge) side 
inference, we achieve a 96.3\% (95.3\%) accuracy in classifying diabetics 
against healthy individuals, and a 95.7\% (94.6\%) accuracy in
distinguishing among type-1 diabetic, type-2 diabetic, and healthy 
individuals. Against conventional baselines, such as support vector
machines with linear and radial basis function kernels, k-nearest
neighbor, random forest, and linear ridge classifiers, DiabNNs achieve 
higher accuracy, while reducing the model size (floating-point operations) 
by up to 454.5$\times$ (8.9$\times$). Therefore, the system can be viewed as 
pervasive and efficient, yet very accurate.
\end{abstract}

}

\maketitle

\IEEEdisplaynontitleabstractindextext

\IEEEpeerreviewmaketitle

\renewcommand{\thefootnote}{}

\IEEEraisesectionheading{\section{Introduction}}
More than 422 million people around the world (more than 24 million in the 
U.S. alone) suffer from diabetes~\cite{_ncd2016worldwide, 
_american2018economic}. This chronic disease imposes a substantial economic 
burden on both the patient and the government, and accounts for nearly 
25\% of the entire healthcare expenditure in the 
U.S.~\cite{_american2018economic}. However, diabetes prevention, care, and 
especially early diagnosis are still fairly challenging given that the disease 
usually develops and gets treated outside a clinic, hence out of reach of 
advanced clinical care. In fact, it is estimated that more than 75\% of the 
patients still remain undiagnosed~\cite{_american2018economic}. This may lead 
to irreversible and costly consequences. For example, studies have shown that 
the longer a person lives with undiagnosed and untreated diabetes, the worse 
their health outcomes are likely to be~\cite{global_report}. Without an early 
alarm, people with pre-diabetes, a less intensive diabetes status that can be 
cured, could end up with diabetes mellitus within five years that can no 
longer be cured~\cite{_rowley2017diabetes}. Thus, it is important to develop an 
accessible and accurate diabetes diagnosis system for the daily life scenario
that can greatly improve general welfare and bend the associated healthcare 
expenditure downwards~\cite{_facchinetti2016continuous}.

The emergence of wearable medical sensors (WMSs) points to a promising way 
to address this challenge. In the past decade, advancements in low-power 
sensors and signal processing techniques have led to many disruptive 
WMSs~\cite{smarthealthcare}. These WMSs 
enable a continuous sensing of physiological signals during 
daily activities, and thus provide a powerful, yet user-transparent, 
human-machine interface for tracking the user's health status. Combining WMSs 
and machine learning brings up the possibility of pervasive health condition 
tracking and disease diagnosis in a daily context~\cite{hdss}. This approach 
exploits the superior knowledge distillation capability of machine
learning to extract medical insights from health-related physiological 
signals~\cite{ml_diagnosis}. Hence, it offers a promising method to bridge 
the information gap that currently separates the clinical and daily domains. 
This helps enable a unified smart healthcare system that serves people in 
both the daily and clinical scenarios~\cite{smarthealthcare}.

However, disease diagnosis based on WMS data and its effective deployment at
the edge still remain challenging~\cite{hdss}. Conventional approaches 
typically involve feature extraction, model training, and model deployment. 
However, such an approach suffers from two major problems:
\begin{itemize}
    \item \textbf{Inefficient feature extraction}: Handcrafting features may 
require substantial engineering effort and expert domain knowledge for
each targeted disease. Searching for informative features through 
trial-and-error can be very inefficient, hence it may not be easy to 
effectively explore the available feature space. This problem is exacerbated 
when the feature space scales up given (i) a growing number of available 
signal types from WMSs, and (ii) more than 69,000 human diseases that
need to be monitored~\cite{icd10}.
    \item \textbf{Vast computation cost}: Due to the need to execute a large number 
of floating-point operations (FLOPs) during feature extraction and model 
inference, continuous health monitoring can be very computationally intensive, 
hence hard to deploy on resource-constrained platforms~\cite{nest}. 
\end{itemize}

To solve these problems, we propose a framework called DiabDeep that combines off-the-shelf WMSs 
with efficient neural networks (NNs) for pervasive diabetes diagnosis. DiabDeep completely bypasses 
the feature extraction stage, acts on raw signals captured by commercially available WMSs, and 
makes accurate diagnostic decisions. It supports inference both on the server and the edge. 
On the resource-rich server, we deploy stacked sparsely connected (SC) layers (DiabNN-server) to 
focus on high accuracy. On the resource-poor edge, we use the hidden-layer long short-term memory 
(H-LSTM) based recurrent layer (DiabNN-edge) to cut down on computation and storage costs while
incurring only a minor accuracy loss. Augmented by a grow-and-prune training methodology, DiabDeep 
simultaneously improves accuracy, shrinks model size, and cuts down on computation costs 
relative to conventional approaches, such as support vector machines (SVMs) and random forest. 

We summarize the major contributions of this article as follows:
\begin{enumerate}
  \item We propose a novel DiabDeep framework that combines off-the-shelf WMSs and efficient NNs 
for pervasive diabetes diagnosis. DiabDeep focuses on both physiological 
and demographic information that can be captured by WMSs in the daily domain, including Galvanic skin response, 
blood volume pulse, inter-beat interval of heart, body temperatures,
ambient environment, body movements, and patient's demographic background.
  \item We design a novel DiabNN architecture that uses different NN layers in its edge and server 
inference model variants to accommodate varying resource budgets and design goals. 
  \item We develop a training flow for DiabNNs based on a grow-and-prune NN synthesis paradigm 
that enables the networks to learn both weights and connections in order to simultaneously tackle 
accuracy and compactness. 
  \item We show that DiabDeep is \textbf{accurate}: we evaluate DiabDeep based on data 
collected from 52 participants. Our system achieves a 96.3\% (95.3\%) accuracy in classifying 
diabetics against healthy individuals on the server (edge), and 95.7\% (94.6\%) accuracy 
in distinguishing among type-1 diabetics, type-2 diabetics, and healthy individuals. 
  \item We show that DiabDeep is \textbf{efficient}: we compare DiabNNs with conventional models, 
including SVMs with linear and radial basis function (RBF) kernels, k-nearest neighbors (k-NN), 
random forest, and linear ridge classifiers. DiabNNs achieve the highest accuracy, while reducing 
model size (FLOPs) by up to 454.5$\times$ (8.9$\times$).
  \item We show that DiabDeep is \textbf{pervasive}: it captures all the signals non-invasively 
through comfortably-worn WMSs that are already commercially available. This greatly assists with 
continuous diabetes detection and monitoring without disrupting daily lifestyle.
\end{enumerate}

The rest of this paper is organized as follows. We review related work in 
Section~\ref{sec:related_work}. Then, in Section~\ref{sec:DiabDeep}, we discuss the proposed DiabDeep framework in detail. We explain our implementation details of DiabDeep in Section~\ref{sec:implementation} and present our experimental results in Section~\ref{sec:exp}. In Section~\ref{sec:discussions}, we discuss the inspirations of our proposed framework from the human brain and future directions inspired by DiabDeep. Finally, we draw conclusions in Section~\ref{sec:conclusion}.

\section{Related Work}
\label{sec:related_work}
In this section, we first discuss diabetes diagnosis approaches using machine learning algorithms 
that have been previously proposed. Then, we focus on recent progress in efficient NN design.

\subsection{Machine learning for diabetes diagnosis}
Numerous studies have focused on applying machine learning algorithms to diabetes diagnosis from 
the clinical domain to the daily scenario.

\noindent
\textbf{Clinical approach:} Electronic health records have been widely used as an information source 
for diabetes prediction and intervention~\cite{_zheng2017machine}. With the recent upsurge in the availability of biomedical datasets, new information 
sources have been unveiled for diabetes diagnosis, including gene 
sequences~\cite{_kavakiotis2017machine} and retinal images~\cite{retina_google}.
However, these approaches are still restricted to the clinical domain, hence have very limited 
access to patient status when he/she leaves the clinic. 

\noindent
\textbf{Daily approach:} 
Daily glucose level detection has recently captured an increasing amount of research attention. One 
stream of study has explored subcutaneous glucose monitoring for continuous glucose tracking in a 
daily scenario~\cite{_facchinetti2016continuous}. This is an invasive approach 
that still requires a high level of compliance, relies on regular sensor replacement (3-14 days), and 
impacts user experience~\cite{_ajjan2019continuous}. Recent systems have started exploiting 
non-invasive WMSs to alleviate these shortcomings. For example, Yin et al. combine machine learning 
ensembles and non-invasive WMSs to achieve a diabetes diagnostic accuracy of 77.6\%~\cite{hdss}. 
Ballinger et al. propose a system called DeepHeart that acts on Apple watch data and patient 
demographics~\cite{deepheart}. DeepHeart uses bidirectional LSTMs to deliver an 84.5\% diagnostic 
accuracy. However, it relies on a small spectrum of WMS signals that include only discrete heart 
rate and step count measurements (indirectly estimated by photoplethysmograph and accelerometer). 
This may lead to information loss, hence reduce diagnostic capability. 
 Swapna et al. achieve a 93.6\% diagnostic accuracy by combining convolutional neural networks (CNNs) 
with LSTMs and heart rate variability measurements~\cite{swapna2018automated}. However, the system 
has to rely on an electroencephalogram (ECG) data stream sampled at 500Hz that is not supported by commercial WMSs. 

\begin{figure*}[t]
\begin{center}
\includegraphics[width=\linewidth]{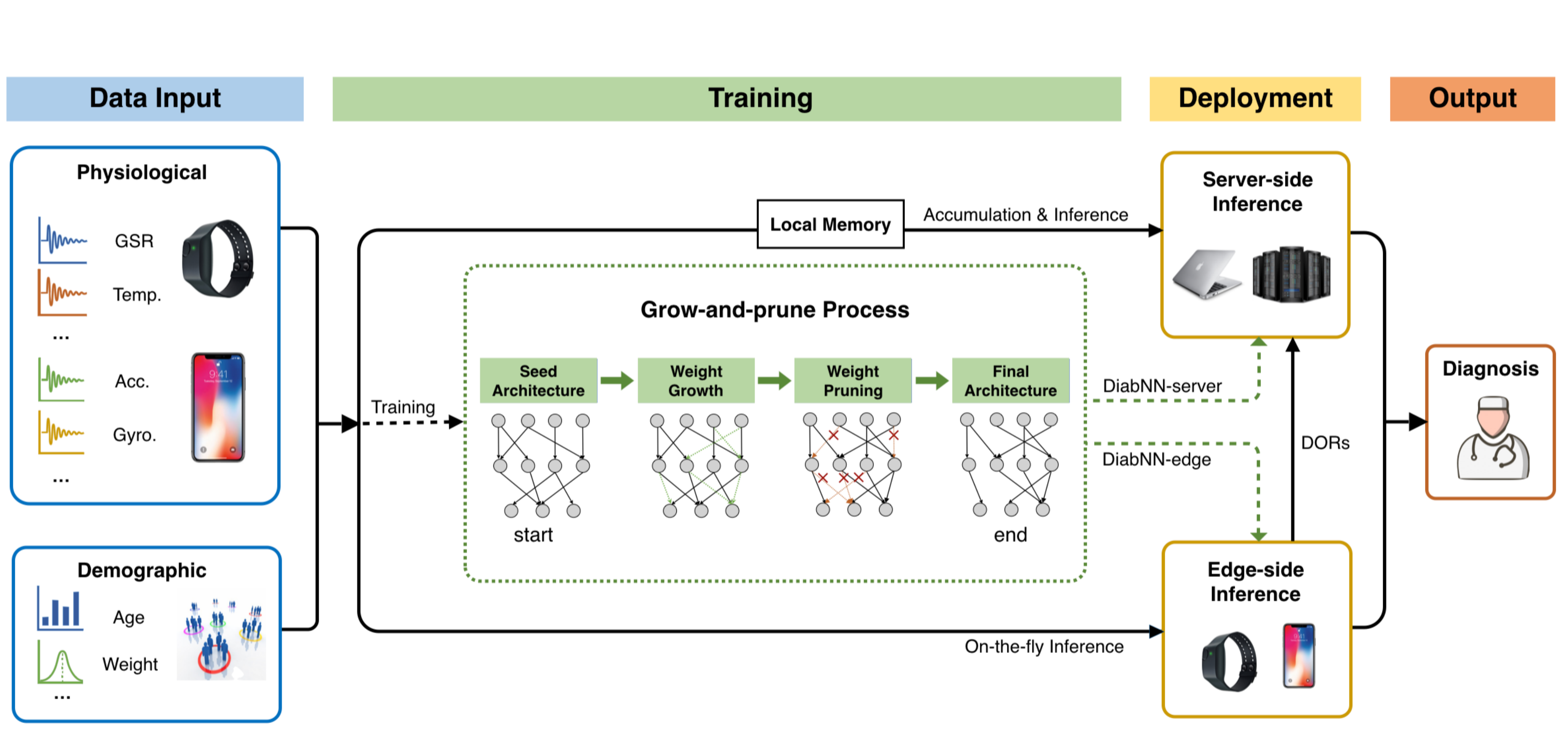}
\end{center}
\caption{The schematic diagram of the proposed DiabDeep framework. DORs refer to disease-onset 
records. GSR, Temp., Acc., and Gyro. refer to Galvanic skin response, skin temperature, accelerometer 
data, and gyroscope data, respectively.}
\label{fig:DiabDeep}
\end{figure*}

\subsection{Efficient neural networks}
Efficient NN design is a vibrant field. We discuss two approaches next.

\noindent
\textbf{Compact model architecture:} One stream of research exploits 
the design of efficient building 
blocks for NN redundancy removal. For example, MobileNetV2 stacks inverted residual building blocks 
to effectively shrink its model size and reduce its FLOPs~\cite{mobilenetv2}. Ma et al. use channel 
shuffle operation and depth-wise convolution to deliver model compactness~\cite{shufflenetv2}. Wu 
et al. propose ShiftNet based on shift-based modules, as opposed to spatial convolution layers, to 
achieve substantial computation and storage cost reduction~\cite{shift}. Besides, automated compact 
architecture design also provides a promising solution~\cite{fbnet, zhou2018neural}. 
Dai et al. develop efficient performance predictors to speed up the search process for efficient 
NNs~\cite{chamnet}. Compared to MobileNetV2 on the ImageNet dataset, the generated ChamNets achieve 
up to 8.5\% absolute top-1 accuracy improvement while reducing inference latency substantially.

\noindent
\textbf{Network compression: } 
Compression techniques~\cite{PruningHS, nest} have emerged as another popular direction for NN 
redundancy removal. The pruning methodology was initially demonstrated to be effective on large 
CNNs by reducing the number of parameters in AlexNet by 9$\times$ and VGG by 13$\times$ for the 
well-known ImageNet dataset, without any accuracy loss~\cite{PruningHS}. Follow-up works have also 
successfully shown its effectiveness on recurrent NNs such as the LSTM~\cite{ese,wenwei,baiduprune}. 
Network growth is a complementary method to pruning that enables a
sparser, yet more accurate, model 
before pruning starts~\cite{nest, incremental}. A grow-and-prune synthesis paradigm typically 
reduces the number of parameters in CNNs~\cite{nest, scann} and LSTMs~\cite{hlstm} by another 
2$\times$, and increases the classification accuracy~\cite{nest}. It enables NN based inference 
even on Internet-of-Things (IoT) sensors \cite{scann}. The model can be further compressed through 
low-bit quantization. For example, Zhu et al. show that a ternary representation of the weights 
instead of full-precision (32-bit) in ResNet-56 can significantly reduce memory cost while
incurring only a minor accuracy loss~\cite{tenary}. The quantized models offer additional speedup 
potential for current NN accelerators~\cite{jia}.

\noindent
\textbf{Knowledge distillation: } Knowledge distillation allows a
compact student network to distill information (or 'dark knowledge')
from a more accurate, but computationally intensive, teacher network (or 
group of teacher networks) by mimicking the prediction distribution, given 
the same data inputs. The idea was first introduced by Hinton et 
al.~\cite{hintonkd}. Since then, knowledge distillation has been
effectively used to discover efficient networks. Romero et al. proposed 
FitNets that distill knowledge from the teacher's hint layers to teach 
compact students~\cite{fitnet}. Passalis et al. enhanced the knowledge 
distillation process by introducing a concept called feature space probability 
distribution loss~\cite{passalis2018learning}. Yim et al. proposed fast 
minimization techniques based on intermediate feature maps that can also 
support transfer learning~\cite{yim2017gift}.

\section{Methodology}
\label{sec:DiabDeep}
In this section, we describe the proposed DiabDeep framework in detail. We first give a high-level 
overview of the entire framework. Then, we zoom into the DiabNN architecture used for DiabDeep 
inference, followed by a detailed description of gradient-based growth and magnitude-based pruning 
algorithms for DiabNN training.

\subsection{The DiabDeep framework}
We illustrate the proposed DiabDeep framework in Fig.~\ref{fig:DiabDeep}. DiabDeep captures both 
physiological and demographic information as data input. It deploys a grow-and-prune training 
paradigm to deliver two inference models, i.e., DiabNN-server and DiabNN-edge, that enable inference 
on the server and on the edge, respectively. Finally, DiabDeep generates diagnosis as output. 
The details of data input, model training, and model inference are as follows:

\begin{itemize}
    \item \textbf{Data input:} As mentioned earlier, DiabDeep focuses on (i) physiological signals 
and (ii) demographic information that are available in the daily context. Physiological signals can 
be captured by WMSs (e.g., from a smartphone and smartwatch) in a non-invasive, passive, and efficient 
manner. The list of collectible signals includes, but is not limited to, heart rate, body 
temperature, Galvanic skin response, and blood volume pulse. Additional signals such as 
eletromechanical and ambient environmental data (e.g., accelerometer, gyroscope, and humidity sensor 
readings) may also provide information on user habit tracking that offers diagnostic 
insights~\cite{hdss}. This list is expanding rapidly, given the speed of ongoing technological 
advancements in this field [17]. Demographics information (e.g., age, weight, gender, and height) 
also assists with disease diagnosis~\cite{hdss}. It can be easily captured and updated through a 
simple user interface on a smartwatch or smartphone. Then, both physiological and demographic data 
are aggregated and merged into a comprehensive data input for subsequent analysis. 
    \item \textbf{Model training:} DiabDeep utilizes a grow-and-prune paradigm to train its NNs, as 
shown in the middle part of Fig.~\ref{fig:DiabDeep}. It starts NN synthesis from a sparse seed 
architecture. It first allows the network to grow connections and neurons based on gradient 
information. Then, it prunes away insignificant connections and neurons based on magnitude 
information to drastically reduce model redundancy. This leads to improved accuracy and 
efficiency~\cite{nest, hlstm}, where the former is highly preferred on the server and the latter 
is critical at the edge. The training process generates two inference models, i.e., DiabNN-server 
and DiabNN-edge, for server and edge inference, respectively. Both models share the same DiabNN 
architecture, but vary in the choice of internal NN layers based on different resource constraints and 
design objectives, as explained later.\\
    \item \textbf{Model inference:} Due to the distinct inference environments encountered upon 
deployment, DiabNN-server and DiabNN-edge require different input data flows, as depicted by 
the separate data paths in Fig.~\ref{fig:DiabDeep}. In DiabNN-server, data have to be accumulated 
in local memory, e.g., local phone/watch storage, before they can be transferred to the 
base station in a daily, weekly, or monthly manner, depending on user preference. As opposed to 
the accumulation-and-inference process, DiabNN-edge enables on-the-fly inference directly at
the edge, e.g., a smartphone. This enables users to receive
instantaneous diagnostic decisions. As 
mentioned earlier, it incurs a slight accuracy degradation (around 1\%) due to the scarce energy 
and memory budgets on the edge. However, this deficit may be alleviated when DiabNN-edge jointly 
works with DiabNN-server. When an alarm is raised, DiabNN-edge can store the relevant data sections 
as disease-onset records (DORs) that can be later transferred to DiabNN-server for further
analysis. In this manner, DiabNN-edge offers a substantial data storage reduction in the
required edge memory by bypassing the storage of 'not-of-interest' signal sections, while preserving 
the capability to make accurate inference on the server side. Such DORs can also be used as 
informative references when future physician intervention and checkup are needed.
\end{itemize}

We next explain our proposed DiabNN architecture in detail.

\begin{figure}[t]
\begin{center}
\includegraphics[width=\linewidth]{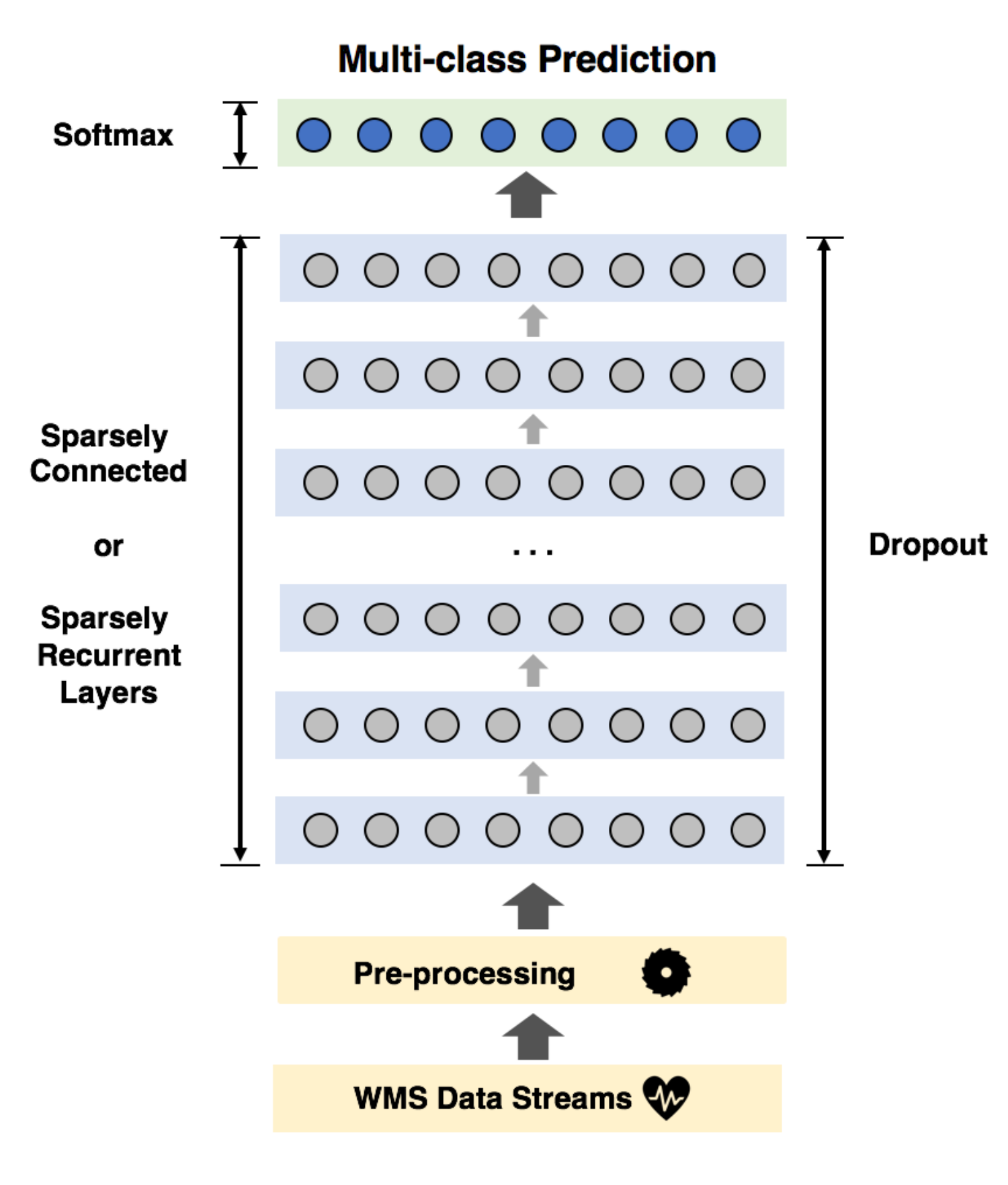}
\end{center}
\caption{An illustration of the DiabNN architecture.}
\label{fig:DiabNN}
\end{figure}

\subsection{The DiabNN architecture}
\label{sec:diabnn}

Fig.~\ref{fig:DiabNN} shows the DiabNN architecture that distills diagnostic decisions (shown
at the top) from data inputs (shown at the bottom). There are three sequential steps employed during 
this process: (i) data preprocessing, (ii) transformation via NN layers, and (iii) output generation 
using softmax. We describe these steps next.

The preprocessing stage is critical for DiabNN inference due to the following 
reasons:
\begin{itemize}
    \item Data normalization: NNs typically favor normalized inputs. Normalization methods, such as 
min-max scaling, standardization, and L2 normalization, generally lead to accuracy and noise 
tolerance improvements~\cite{alexnet, sensitivity_analysis}. In this work, we apply min-max 
scaling to scale each input data stream into the [0,1] range:
    \begin{align*}
    \textbf{x}_{normalized} = \frac{\textbf{x} - min(\textbf{x})}{max(\textbf{x}) - min(\textbf{x})} 
    \end{align*}
    \item Data alignment: WMS data streams may vary in their start times and sampling
frequencies~\cite{him}. Therefore, we guarantee that the data streams are synchronized by checking 
their timestamps and applying appropriate offsets accordingly.
\end{itemize}

We use different NN layers in DiabNN for server and edge inference. DiabNN-server deploys SC layers 
to aim at high accuracy whereas DiabNN-edge utilizes sparsely recurrent (SR) layers to aim at 
extreme efficiency. All NN layers are subjected to dropout regularization, which is a widely-used 
approach for addressing overfitting and improving accuracy~\cite{dropout}. 

In DiabNN-server, each SC layer conducts a linear transformation (using a sparse matrix as opposed 
to a conventional full matrix) followed by a nonlinear activation function. As shown later, utilizing 
SC layers leads to more model parameters than SR layers, hence leads to an improved learning 
capability and higher accuracy. Consequentially, DiabNN-server achieves a 1$\%$ accuracy 
improvement over DiabNN-edge.

\begin{figure}[t]
\begin{center}
\includegraphics[width=\columnwidth]{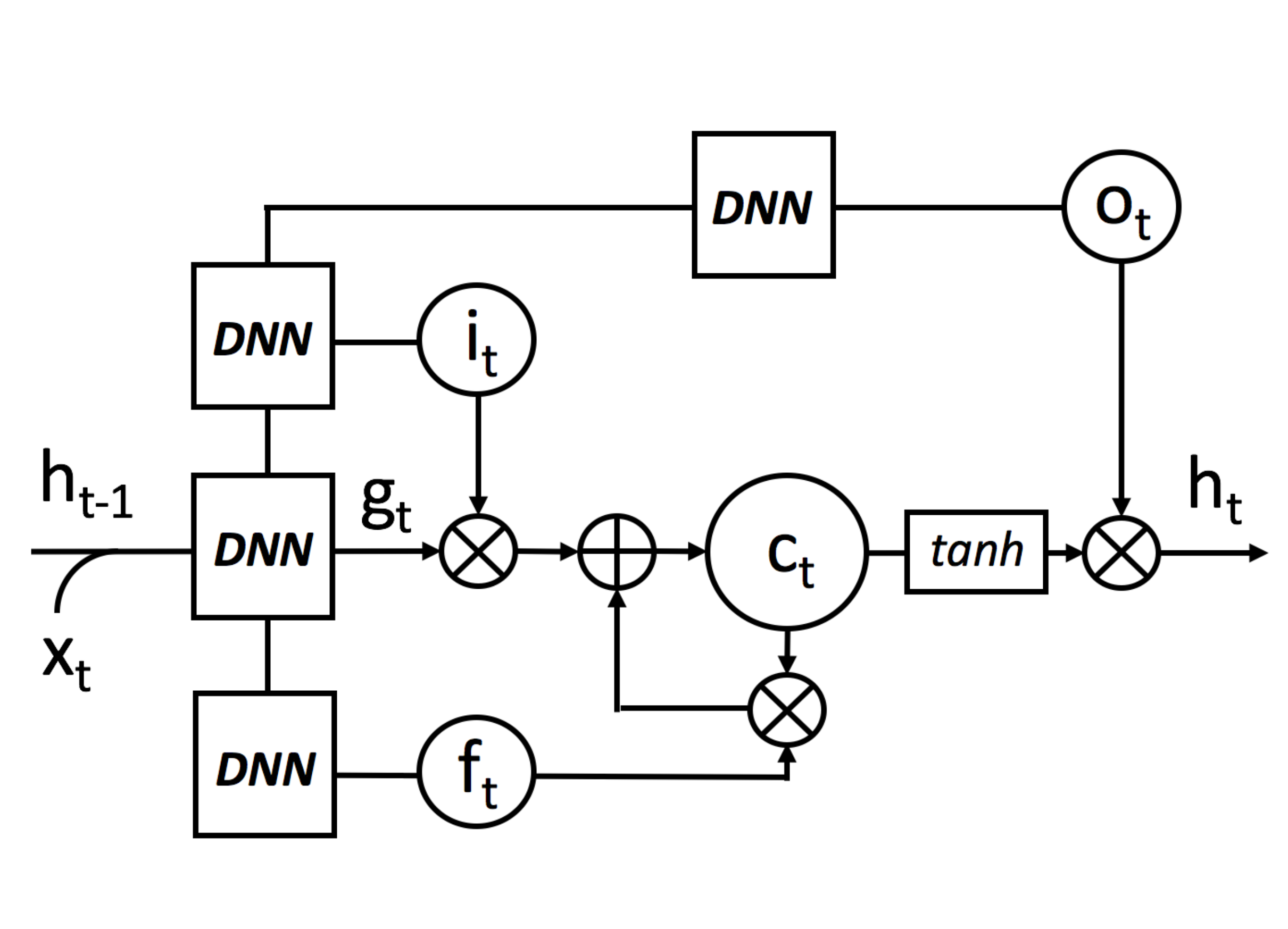}
\end{center}
\caption{The schematic diagram of an H-LSTM cell. DNNs refer to deep neural network control gates.}
\label{fig:hlstm}
\end{figure}

In DiabNN-edge, we base our SR layer design on the H-LSTM cell \cite{hlstm}. It is a variant of the 
conventional LSTM cell obtained through addition of hidden layers to its control gates. 
Fig.~\ref{fig:hlstm} shows the schematic diagram of an H-LSTM. Its internal computation flow 
is governed by the following equations: 

\begin{equation*}
\begin{split}
\textbf{f}_{t} &= \sigma(\textbf{W}^s_{f} H^{*}([\textbf{x}_{t},\textbf{h}_{t-1}])+\textbf{b}_f)\\
\textbf{i}_{t} &= \sigma(\textbf{W}^s_i H^{*}( [\textbf{x}_{t},\textbf{h}_{t-1}])+\textbf{b}_o)\\
\textbf{o}_{t} &= \sigma(\textbf{W}^s_o H^{*}( [\textbf{x}_{t},\textbf{h}_{t-1}])+\textbf{b}_o) \\
\textbf{g}_{t} &= tanh(\textbf{W}^s_g H^{*}( [\textbf{x}_{t},\textbf{h}_{t-1}])+\textbf{b}_g) \\
\textbf{c}_{t} &= \textbf{f}_{t} \otimes \textbf{c}_{t-1} + \textbf{i}_{t} \otimes \textbf{g}_{t}\\
\textbf{h}_{t} &= \textbf{o}_{t} \otimes tanh(\textbf{c}_t)\\
\end{split}
\end{equation*}

\noindent
where $\textbf{f}_{t}$, $\textbf{i}_{t}$, $\textbf{o}_{t}$, $\textbf{g}_{t}$, $\textbf{x}_{t}$, 
$\textbf{h}_{t}$, and $\textbf{c}_{t}$ denote the forget gate, input gate, output gate, cell update 
vector, input, hidden state, and cell state at step $t$, respectively; $\textbf{h}_{t-1}$ and 
$\textbf{c}_{t-1}$ refer to the previous hidden and cell states at step $t-1$; $H$, $\textbf{W}^{s}$, 
$\textbf{b}$, $\sigma$, and $\otimes$ refer to a hidden layer that performs a linear transformation 
followed by an activation function, sparse weight matrix, bias, $sigmoid$ function, and element-wise 
multiplication, respectively; $^{*}$ indicates zero or more $H$ layers for each NN gate. The 
additional hidden layers enable three advantages. First, they enhance gate control through a 
multi-level abstraction that can lead to accuracy gains. Second, they can be easily regularized 
through dropout, and thus lead to better generalization. Third, they offer a wide range of choices 
for internal activation functions, such as the rectified linear unit (ReLU), that can lead to faster 
learning~\cite{hlstm}. Using H-LSTM based SR layers, DiabNN-edge reduces the model size by 
130$\times$ and inference FLOPs by 2.2$\times$ relative to DiabNN-server. 

\begin{algorithm}[tb]
\SetAlgoLined
   \caption{Gradient-based growth}
   \label{alg:growth}
   {\bfseries Input:} $\textbf{W} \in R^{M\times N}$: weight matrix of dimension
$M\times N$, $\textbf{Msk}\in R^{M\times N}$: weight mask of dimension $M\times N$ \\
   {\bfseries Output:} updated $\textbf{Msk}$, updated $\textbf{W}$   \\
   {\bfseries Denote:} $\alpha$: growth ratio, $\textbf{W}.grad$: gradient of the weight matrix, 
   $\eta $: current learning rate\\
   \textbf{Begin}\\
   Accumulate $\textbf{W}.grad$ for one training epoch \\
   $thres$ = $(\alpha MN)^{th}$ largest element in $ \vert \textbf{W}.grad  \vert$\\
   \For {$1\leq m\leq M$}{
    \For {$1\leq n\leq N$}{
       \If {$|W.grad_{m, n}| > thres$}{
          $Msk_{m, n} = 1 $}
     }
    }
    $\textbf{W}$ $\leftarrow \textbf{W} + (\eta \times \textbf{W.grad}  \otimes \textbf{Msk}$)\\
   {\bfseries Return} $\textbf{Msk}$, $\textbf{W}$\\
   \textbf{End}
\end{algorithm}

\begin{algorithm}[tb]
\SetAlgoLined
   \caption{Magnitude-based pruning}
   \label{alg:pruning}
   {\bfseries Input:} $\textbf{W} \in R^{M\times N}$: weight matrix of dimension $M\times N$, $\textbf{Msk}\in R^{M\times N}$: weight mask of dimension $M\times N$ \\
   {\bfseries Denote:} $\beta$: pruning ratio\\
   {\bfseries Output:} updated $\textbf{Msk}$, updated $\textbf{W}$  \\
   \textbf{Begin}\\
   $thres$ = $(\beta MN)^{th}$ largest element in $ \vert \textbf{W}  \vert$ \\
   \For {$1\leq m\leq M$}{
   \For{$1\leq n\leq N$}{
   \If {$|W_{m, n}| < thres$}{
   $Msk_{m, n} = 0 $
   }
   }
   }
   $\textbf{W} \leftarrow \textbf{W}  \otimes \textbf{Msk}$\\
   {\bfseries Return} $\textbf{Msk}$, $\textbf{W}$\\
   \textbf{End}
\end{algorithm}

\subsection{Grow-and-prune training for DiabNN}

We next explain the gradient-based network growth and magnitude-based network pruning algorithms in 
detail. Unless otherwise stated, we assume a mask-based approach for tackling sparse networks. Each 
weight matrix \textbf{W} has a corresponding binary mask matrix \textbf{Msk} that has the exact same 
size. It is used to disregard dormant connections (connections with zero-valued weights).

Algorithm~\ref{alg:growth} illustrates the connection growth process. The main objective of the 
weight growth phase is to locate only the most effective dormant connections to reduce the value 
of the loss function $L$. To do so, we first evaluate the gradient for all the dormant connections 
and use this information as a metric for ranking their effectiveness. During the training process, 
we extract the gradient of all weight matrices ($\textbf{W}.grad$) for each mini-batch of training 
data using the back-propagation algorithm. We repeat this process over a whole training epoch to 
accumulate $\textbf{W}.grad$. Then, we calculate the average gradient over the entire epoch by 
dividing the accumulated values by the number of training instances. We activate a dormant 
connection $w$ if and only if its gradient magnitude is larger than the $((1-\alpha)\times100)^{th}$ 
percentile of the gradient magnitudes of its associated layer matrix. Its initial value is set
to the product of its gradient value and the current learning rate. The growth ratio $\alpha$ is a 
hyperparameter. We typically use $0.1\leq \alpha \leq 0.3$ in our experiments. The NN growth method 
was first proposed in~\cite{nest}. It has been shown to be very effective in enabling the network 
to reach a higher accuracy with far less redundancy than a fully connected model.

We show the connection pruning algorithm in Algorithm~\ref{alg:pruning}. During this process, we 
remove a connection $w$ if and only if its magnitude is smaller than the 
$(\beta\times100) ^ {th}$ percentile of the weight magnitudes of its associated layer matrix. When 
pruned away, the connection's weight value and its corresponding mask binary value are simultaneously 
set to zero. The pruning ratio $\beta$ is also a hyperparameter. Typically, we use 
$\beta \leq 0.3$ in our experiments. Connection pruning is an iterative process, where we retrain 
the network to recover its accuracy after each pruning iteration.

\begin{figure}[t]
\begin{center}
\includegraphics[width=\columnwidth]{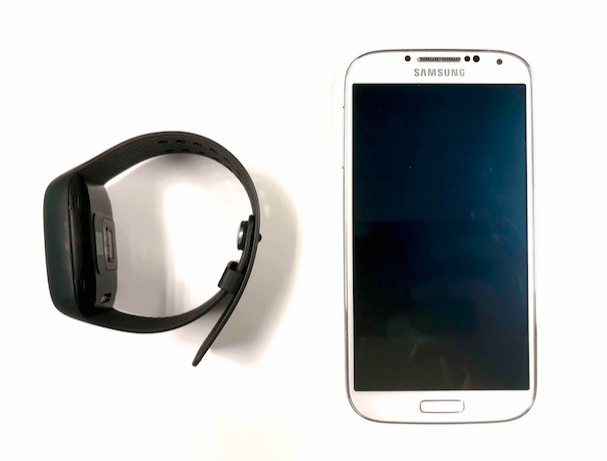}
\end{center}
\caption{A photo of the Empatica E4 smartwatch (left) and the Samsung Galaxy S4 smartphone (right) used for data collection.}
\label{fig:wms}
\end{figure}

\begin{table}[ht]
\caption{Data types collected from each participant}\label{tb:1}
\centering 
\begin{tabular}{lc}
\hline
\hline
\textbf{Data type} & \textbf{Source}\\
\hline
Galvanic skin response & Smart watch\\
Skin temperature        & Smart watch\\
Acceleration ($x, y, z$) & Smart watch\\
Inter-beat interval & Smart watch\\
Blood volume pulse & Smart watch\\
\hline
Humidity & Smart phone \\
Ambient illuminance & Smart phone \\
Ambient light color spectrum & Smart phone \\
Ambient temperature  & Smart phone \\
Gravity ($x, y, z$) & Smart phone \\
Angular velocity ($x, y, z$) & Smart phone \\
Orientation ($x, y, z$) & Smart phone \\
Acceleration ($x, y, z$)   & Smart phone \\
Linear acceleration ($x, y, z$) & Smart phone \\
Air pressure & Smart phone \\
Proximity & Smart phone \\
Wi-Fi radiation strength & Smart phone \\
Magnetic field strength & Smart phone \\
\hline
Age & Questionnaire \\
Gender & Questionnaire \\
Height & Questionnaire \\
Weight & Questionnaire \\
Relatives with diabetes & Questionnaire \\
Smoking & Questionnaire \\
Drinking & Questionnaire \\
\hline
\hline
\label{tb:signal}
\end{tabular}
\end{table}

\section{Implementation Details}
\label{sec:implementation}

In what follows, we first describe the dataset collected from 52 participants that is used for 
DiabDeep evaluation. Then, we explain the implementation details of DiabDeep based on the collected
dataset.

\subsection{Data collection and preparation}

In this study, we collected both the physiological data and demographic information from 52 
participants. 27 participants were diagnosed with diabetes (14 with type-1 and 13 
with type-2 diabetes) whereas the remaining 25 participants were healthy non-diabetic baselines. 
We collected the physiological data using a commercially available Empatica E4 
smartwatch \cite{e4_connect} and Samsung Galaxy S4 smartphone, as shown in Fig.~\ref{fig:wms}. We also 
used a questionnaire to gather demographic information from all the participants. We summarize all 
the data types collected in this study in Table~\ref{tb:signal}. It can be observed that the
collected data cover a wide range of physiological and demographic signals 
that may assist with diabetes diagnosis in the daily context. The smartwatch 
data capture the physiological state of the target user. This information, 
e.g., GSR (measures the electrical activity of the skin, i.e., skin 
conductance) and BVP (measures cardiovascular activity, e.g., heart beat 
waveform and heart rate variability, etc.), has been shown to
effectively capture the body status in terms of its health 
indicators~\cite{smarthealthcare}. The ambient information from the smartphone 
may assist with sensing of body movement and physiological signal calibration. 
Finally, demographic information has been previously shown to be effective 
for diabetes diagnosis~\cite{hdss}. In this work, we study whether synergies 
among sources of the above information collected in the daily context can 
support the task of pervasive diabetes diagnosis.

During data collection, we first inform all the participants about the experiment, let them sign 
the consent form, and ask them to fill the demographic questionnaire. Then, we place the Empatica E4 
smartwatch on the wrist of participant's non-dominant hand, and the Samsung Galaxy S4 smartphone in 
the participant's pocket. The experiment lasts between 1.0 and 1.5 hours
per participant during which time the smartwatch and smartphone continuously 
track and store the physiological signals. 
We use the Empatica E4 Connect portal for smartwatch data retrieval~\cite{e4_connect}. 
We develop an Android application to record 
all the smartphone sensor data streams. All the data streams contain detailed timestamps that are 
later used for data synchronization. The experimental procedure was approved by the Institutional 
Review Board of Princeton University. None of the participants reported mental, cardiac, or endocrine 
disorders.

We next preprocess the dataset before training the model. We first synchronize and window the
WMS data streams. To avoid time correlation between adjacent data windows, we divide data into 
$t = 15s$ windows with $s = 30s$ shifts in between. The final dataset contains 5030 data instances. 
We use 70\%, 10\%, and 20\% of the data as training, validation, and test sets. The training, 
validation, and test sets have no time overlap. We then extract the value ranges of the data 
streams from the training set, and then scale all three datasets based on the min-max scaling 
method, as explained earlier.

\subsection{DiabDeep implementation}

We implement the DiabDeep framework using PyTorch~\cite{pytorch} on Nvidia GeForce GTX 1060 GPU 
(with 1.708GHz frequency and 6GB memory) and Tesla P100 GPU (with 1.329GHz frequency and 16GB 
memory). We employ CUDA 8.0 and CUDNN 5.1 libraries in our experiments. We next describe our implementation of DiabNNs based on the collected dataset.

\subsubsection{DiabNN-server}

\begin{figure}[t]
\begin{center}
\includegraphics[width=5cm]{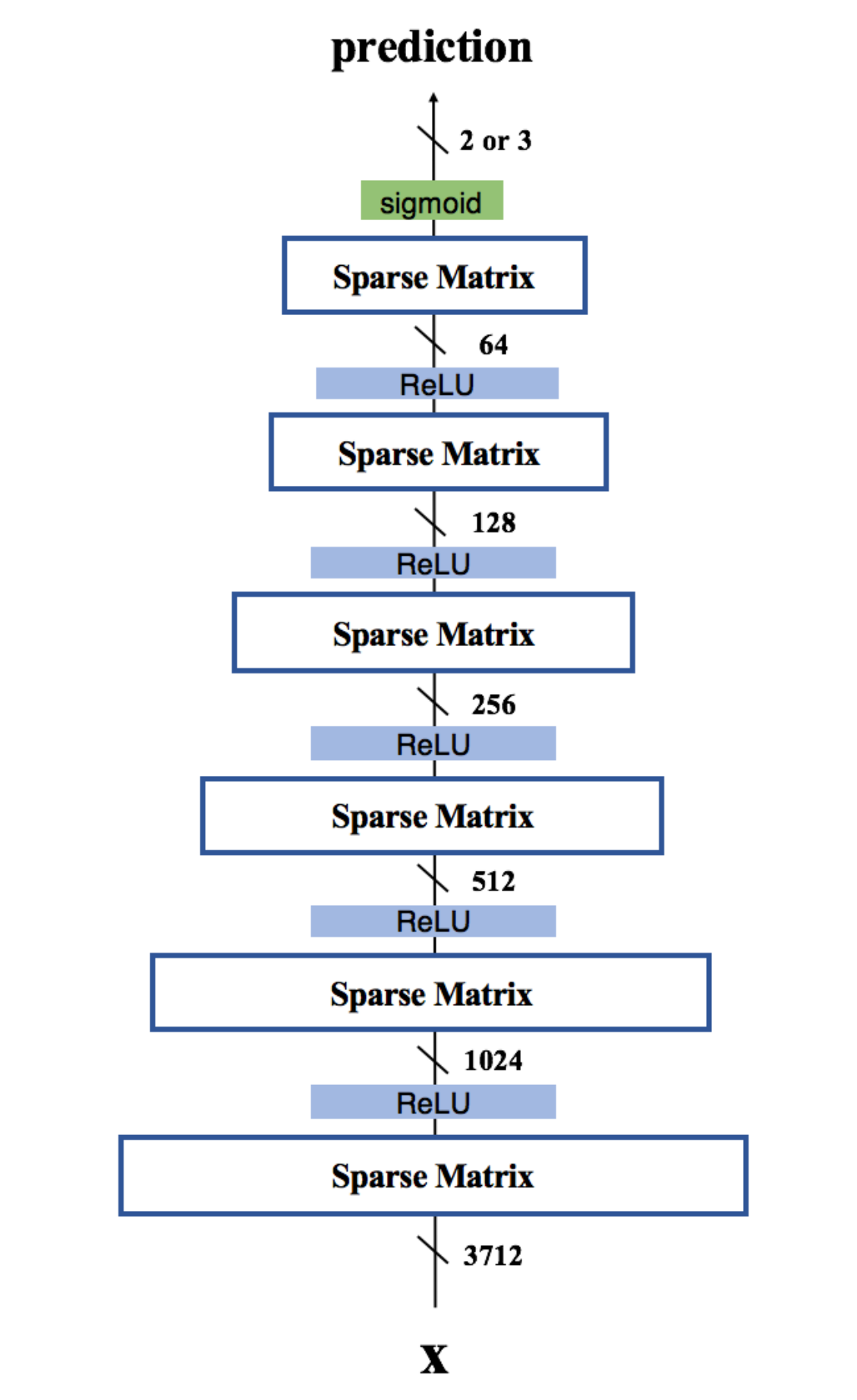}
\end{center}
\caption{Architecture of DiabNN-server. \textbf{x} denotes the input tensor.}
\label{fig:diabnn-server}
\end{figure}

\begin{figure}[t]
\begin{center}
\includegraphics[width=6cm]{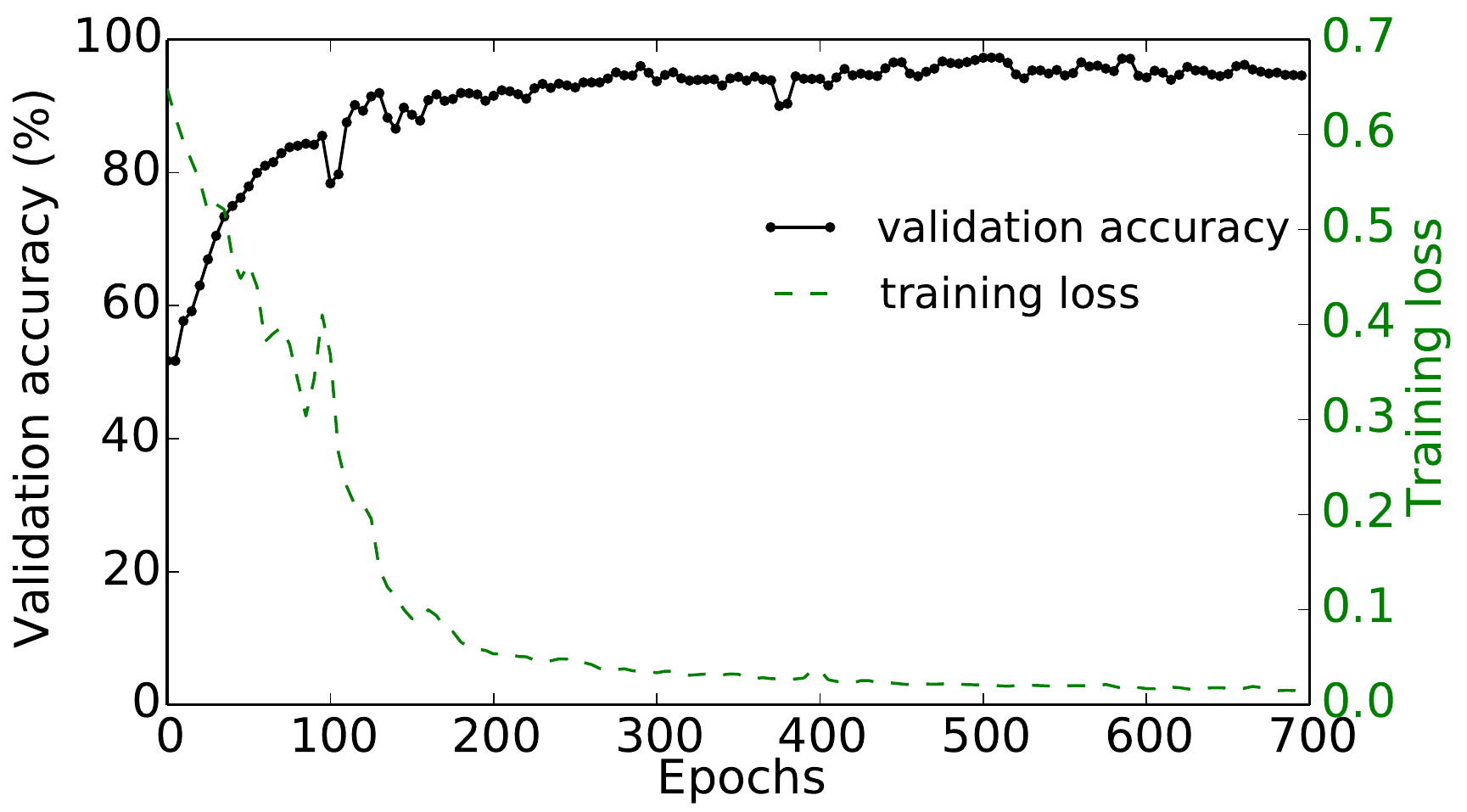}
\end{center}
\caption{Training curve for DiabNN-server to distinguish among type-1 diabetic, type-2 diabetic, and healthy individuals.}
\label{fig:validation_server}
\end{figure}

We first explain the implementation details for DiabNN-server.

\textbf{Data input:} For each data instance, we flatten and concatenate the data within the same 
monitoring window from both the smartphone and smartwatch. This results in a vector of length 3705,
where the flattened smartwatch window contains 2535 signal readings 
(from one data stream at 64Hz, three data streams at 32Hz, two data streams at 4Hz and one data stream at 1Hz), 
and the flattened smartphone window provides additional 1170 signal readings 
(from 26 data streams at 3Hz). Finally, we append the seven demographic
features at its end and obtain a vector of length 3712 as the input for DiabNN-server. 

\textbf{Model architecture:} We present the model architecture for DiabNN-server in Fig.~\ref{fig:diabnn-server}.
We use six sequential SC layers in DiabNN-server with widths set at 
1024, 512, 256, 128, 64 and 2 (3 for three-class classification), respectively. 
The input dimension is the same as the input tensor 
dimension of 3712. We use ReLU as the nonlinear activation function for all SC layers. 

\textbf{Training:} We use a stochastic gradient descent (SGD) optimizer with a momentum of 0.9 for 
this experiment. We initialize the learning rate to 0.005 and divide the learning rate by 10 when 
the validation accuracy does not increase in 50 consecutive epochs. We use a batch size of 256 and a 
dropout ratio of 0.2. For grow-and-prune training, we initialize the seed architecture with a filling 
rate of 20\%. We grow the network for three epochs using a 0.2 growth ratio. For network pruning, we 
initialize the pruning ratio to 0.2. We halve the pruning ratio if the retrained model cannot restore 
accuracy on the validation set. We terminate the process when the ratio falls below 0.01. The
training curve for DiabNN-server is presented in Fig.~\ref{fig:validation_server}.

\begin{figure}[t]
\begin{center}
\includegraphics[width=\columnwidth]{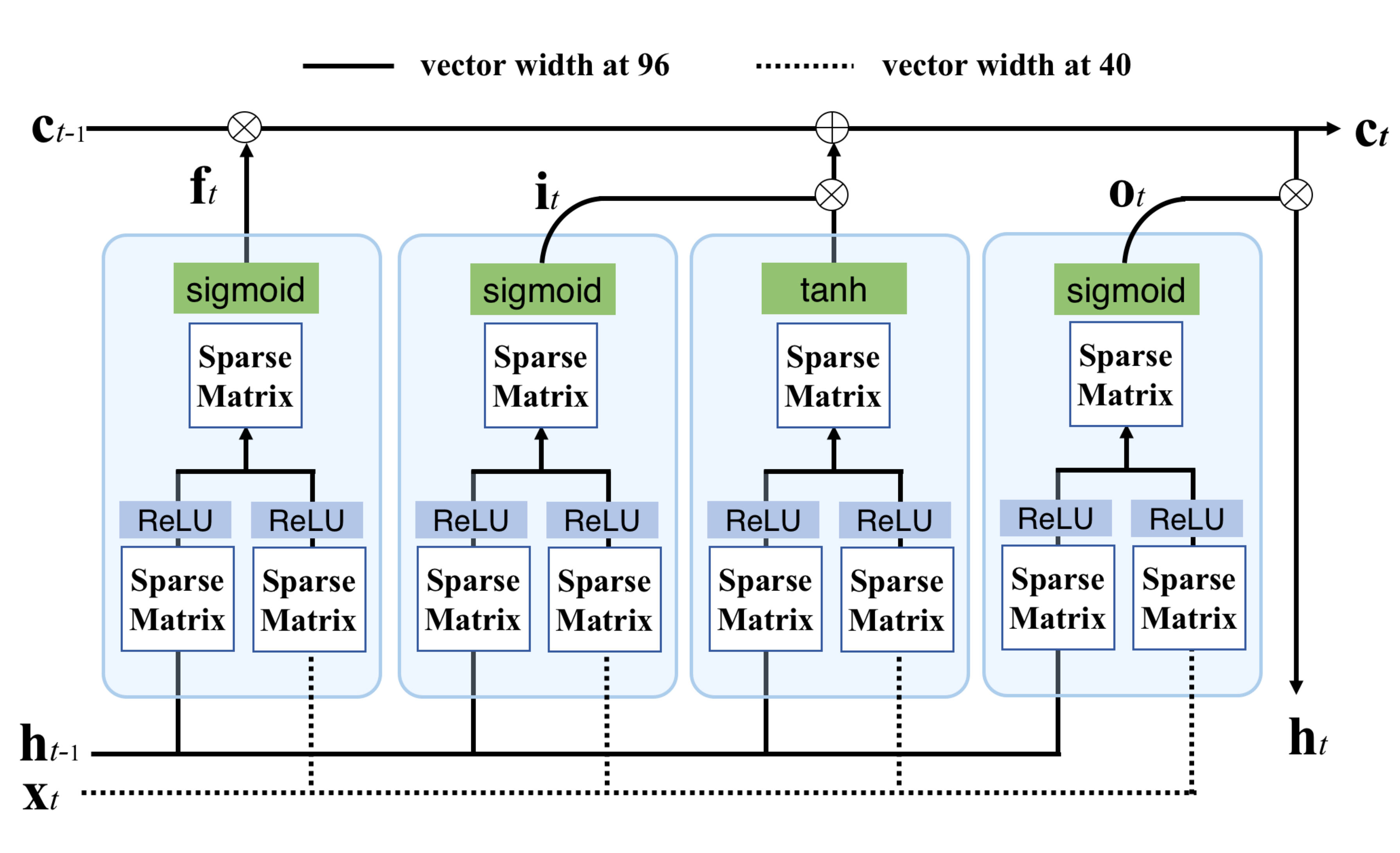}
\end{center}
\caption{Architecture of DiabNN-edge. The denotations follow the description in Section~\ref{sec:diabnn}.}
\label{fig:diabnn-edge}
\end{figure}

\begin{figure}[t]
\begin{center}
\includegraphics[width=6cm]{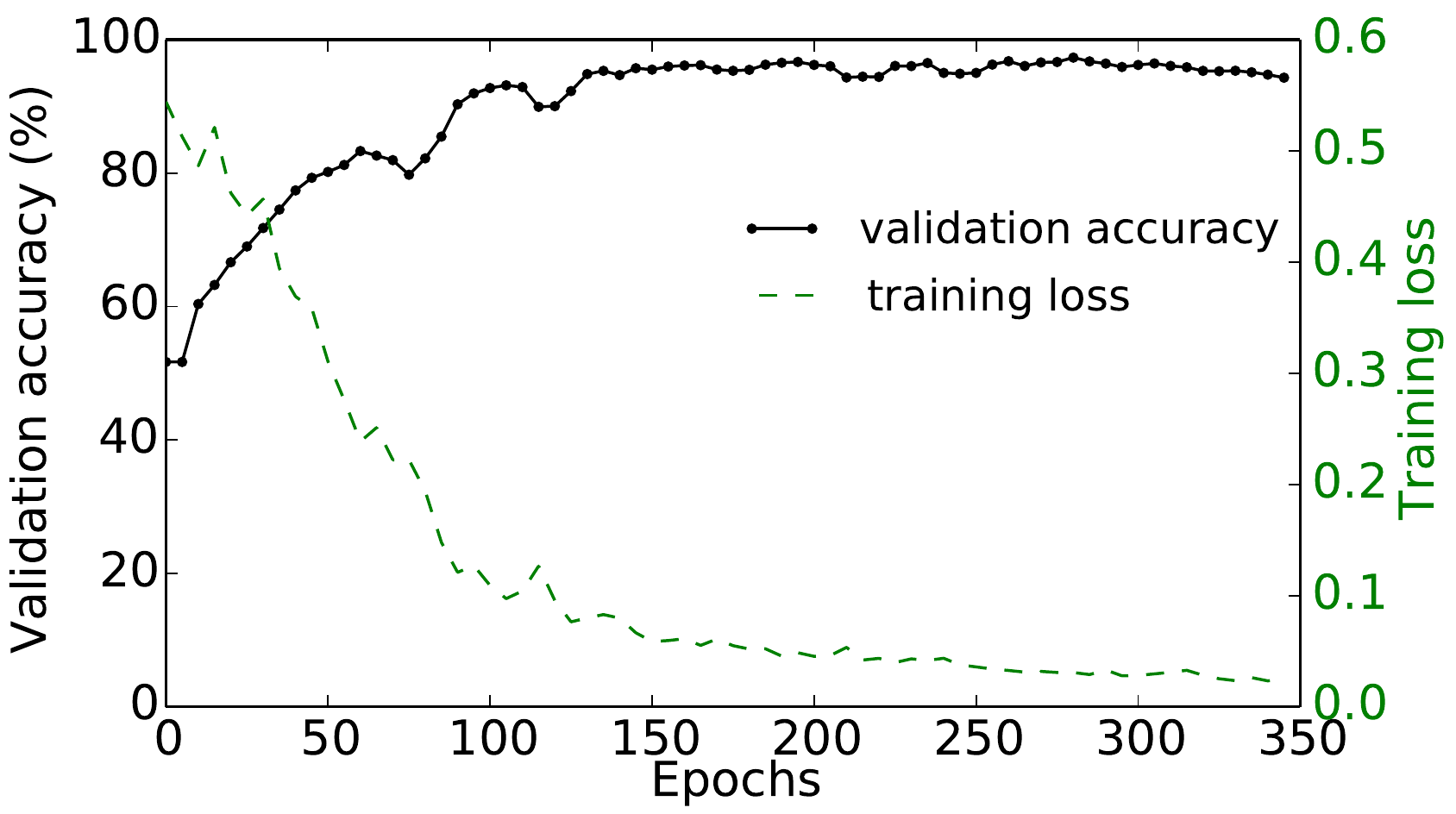}
\end{center}
\caption{Training curve for DiabNN-edge to distinguish among type-1 diabetic, type-2 diabetic, and healthy individuals.}
\label{fig:validation_edge}
\end{figure}

\subsubsection{DiabNN-edge}

We explain the implementation details for DiabNN-edge in this section.

\textbf{Data input:} 
Unlike SC layer based DiabNN-server, SR layer based DiabNN-edge acts on time series data step by 
step~\cite{hlstm}. Thus, at each time step, we concatenate the temporal signal values from each 
data stream along with the demographic information to form an input vector of length 40
(corresponding to seven smartwatch data streams, 26 smartphone data streams, 
and seven demographic features, as shown in Table~\ref{tb:1}). DiabNN-edge operates on four input vectors per second. 
When a signal reading is missing in a data stream 
(e.g., due to a lower sampling frequency), we use the closest previous reading in that data stream 
as the interpolated value.

\textbf{Model architecture:} We present the model architecture for DiabNN-edge in Fig.~\ref{fig:diabnn-edge}. 
DiabNN-edge contains one H-LSTM cell based SR layer that has a hidden 
state width of 96. Each control gate within the H-LSTM cell contains one hidden layer. We use ReLU as 
the nonlinear activation function. 

\textbf{Training:} We again use an SGD optimizer with a momentum of 0.9 for this experiment. The 
learning rate is initialized to 0.001. We divide the learning rate by 10 when the validation accuracy 
does not increase in 30 consecutive epochs. We use a batch size of 64 and a dropout ratio of 0.2 for 
training. For grow-and-prune training, we use the same hyperparameter set as in the experiment for 
DiabNN-server. The training curve for DiabNN-edge is presented in Fig.~\ref{fig:validation_edge}.

\begin{table}[t]
\caption{Performance evaluation metrics}
\centering
\resizebox{8cm}{!}{%
\begin{tabular}{cc}
\hline
\hline
\textbf{Name}&\textbf{Definition}\\
\hline
Accuracy &$(TP+TN) / (TP+FP+FN+TN)$\\
False positive rate &$FP / (TN+FP)$\\
False negative rate &$FN / (TP+FN)$\\
F1 score & $2TP/(2TP+FP+FN)$\\
\hline
\hline
\multicolumn{2}{l}{\scriptsize{$TP (TN)$: diabetic (healthy) instances classified as diabetic (healthy)}}\\
\multicolumn{2}{l}{\scriptsize{$FP (FN)$: healthy (diabetic) instances classified as diabetic (healthy)}}\\
\end{tabular}%
}\\
\label{tb:performance-matrix}
\end{table}

\section{Evaluating DiabDeep Performance}
\label{sec:exp}

In this section, we first analyze the performance of DiabNN-server and DiabNN-edge for two 
classification tasks: (i) binary classification that distinguishes between diabetic vs. healthy 
individuals, and (ii) three-class classification to distinguish among type-1 diabetic, type-2 
diabetic, and healthy individuals. Then, we compare the performances of 
DiabNN-server, DiabNN-edge, and the relevant baselines.

We evaluate the performance of DiabNNs using four performance metrics, as 
summarized in Table~\ref{tb:performance-matrix}. Accuracy indicates the overall prediction capability. 
The false positive rate (FPR) and false negative rate (FNR) measure the 
DiabNN's capability to avoid misclassifying healthy and diabetic instances, 
respectively. The F1 score measures the overall performance
of precision and sensitivity.

\subsection{DiabNN-server performance evaluation}

We first analyze the performance of DiabNN-server. Table~\ref{tb:conf_DiabNN_server_2class} 
presents the confusion matrix of DiabNN-server for the 
binary classification task. DiabNN-server achieves an overall accuracy of 96.3\%. For the healthy 
instances, it achieves a very low FPR of 4.3\%, demonstrating its effectiveness in avoiding false
alarms. For the diabetic instances, it achieves an FNR of 3.1\%, indicating its effectiveness in
raising alarms when diabetes does occur. DiabNN-server achieves an F1 score of 
96.5\% for the binary classification task.

We present the confusion matrix of DiabNN-server for the three-class classification task in 
Table~\ref{tb:conf_DiabNN_server_3class}. DiabNN-server achieves an overall accuracy of 95.7\%. 
For the healthy instances, it achieves a low FPR of 6.6\%, again demonstrating its ability to
avoid false alarms. It also delivers low FNRs for both type-1 and type-2 diabetic individuals of
1.6\% and 2.8\%, respectively (each FNR depicts the ratio of the number of false predictions for a 
target diabetes type divided by the total number of instances of that type). DiabNN-server 
achieves an F1 score of 95.7\% for the three-class classification task.

Furthermore, the grow-and-prune training paradigm not only delivers high diagnostic accuracy, but 
also leads to model compactness as a side benefit. For binary classification, the final 
DiabNN-server model contains only 429.1K parameters with a sparsity level of 90.5\%.
For the three-class classification task, the final DiabNN-server model contains only 445.8K 
parameters with a sparsity level of 90.1\%. The model compactness achieved in both cases 
can help reduce storage and energy consumption on the server.

\subsection{DiabNN-edge performance evaluation}

We next analyze the performance of DiabNN-edge. We present the confusion matrix of DiabNN-edge 
for the binary classification task in 
Table~\ref{tb:conf_DiabNN_edge_2class}. DiabNN-edge achieves an overall accuracy of 95.3\%. For the 
healthy case, it also achieves a very low FPR of 3.7\%. For diabetic instances, it achieves an FNR 
of 5.6\%. This shows that DiabNN-edge can also effectively raise disease
alarms on the edge. DiabNN-edge achieves an F1 score of 95.4\% for the binary
classification task.

\renewcommand{\arraystretch}{2.7}
\noindent
\begin{center}
\begin{table}
\label{tb:conf_DiabNN_server}
\caption{DiabNN-server confusion matrix for binary classification}
\begin{tabular}{l|l|c|c|c}
\multicolumn{2}{c}{}&\multicolumn{2}{c}{prediction}&\\
\cline{3-4}
\multicolumn{2}{c|}{}& diabetic & healthy &\multicolumn{1}{c}{total}\\
\cline{2-4}
\multirow{2}{*}{\qquad \qquad label}& diabetic & 504  & 16 & 520\\
\cline{2-4}
& healthy & 21 & 465 & 486 \\
\cline{2-4}
\multicolumn{1}{c}{} & \multicolumn{1}{c}{total} & \multicolumn{1}{c}{525} & \multicolumn{1}{c}{481} & \multicolumn{1}{c}{1006}\\
\end{tabular}
\label{tb:conf_DiabNN_server_2class}
\end{table}
\end{center}
\renewcommand{\arraystretch}{1}

\renewcommand{\arraystretch}{2.5}
\noindent
\begin{table}
\caption{DiabNN-server confusion matrix for three-class classification}
\begin{tabular}{l|l|c|c|c|c}
\multicolumn{2}{c}{}&\multicolumn{3}{c}{prediction}\\
\cline{3-5}
\multicolumn{2}{c|}{} & type-1 & type-2 & healthy &  \multicolumn{1}{c}{total}\\
\cline{2-5}
\multirow{3}{*}{\qquad label}          & type-1 & 303              & 1   & 4   &  308\\
\cline{2-5}
                                & type-2 & 5                & 206    & 1 & 212 \\
\cline{2-5}
                                & healthy & 22             &  10    & 454 & 486 \\
\cline{2-5}
\multicolumn{1}{c}{} & \multicolumn{1}{c}{total} & \multicolumn{1}{c}{330} & \multicolumn{1}{c}{217} & \multicolumn{1}{c}{459} & {1006}\\
\end{tabular}
\label{tb:conf_DiabNN_server_3class}
\end{table}
\renewcommand{\arraystretch}{1}

\renewcommand{\arraystretch}{2.7}
\noindent
\begin{table}[h]
\caption{DiabNN-edge confusion matrix for binary classification}
\begin{tabular}{l|l|c|c|c}
\multicolumn{2}{c}{}&\multicolumn{2}{c}{prediction}&\\
\cline{3-4}
\multicolumn{2}{c|}{}& diabetic & healthy &\multicolumn{1}{c}{total}\\
\cline{2-4}
\multirow{2}{*}{ \qquad \qquad label}& diabetic &  491 & 29 & 520\\
\cline{2-4}
& healthy & 18 & 468 & 486 \\
\cline{2-4}
\multicolumn{1}{c}{} & \multicolumn{1}{c}{total} & \multicolumn{1}{c}{509} & \multicolumn{1}{c}{497} & \multicolumn{1}{c}{1006}\\
\end{tabular}
\label{tb:conf_DiabNN_edge_2class}
\end{table}

\noindent
\begin{table}
\caption{DiabNN-edge confusion matrix for three-class classification}
\begin{tabular}{l|l|c|c|c|c}
\multicolumn{2}{c}{}&\multicolumn{3}{c}{prediction}\\
\cline{3-5}
\multicolumn{2}{c|}{} & type-1 & type-2 & healthy &  \multicolumn{1}{c}{total}\\
\cline{2-5}
\multirow{3}{*}{\qquad label}          & type-1 & 288              & 4   & 16   & 308 \\
\cline{2-5}
                                & type-2 & 7                & 200    & 5 & 212 \\
\cline{2-5}
                                & healthy & 12             &  10    & 464 & 486 \\
\cline{2-5}
\multicolumn{1}{c}{} & \multicolumn{1}{c}{total} & \multicolumn{1}{c}{307} & \multicolumn{1}{c}{214} & \multicolumn{1}{c}{485} & {1006}\\
\end{tabular}
\label{tb:conf_DiabNN_edge_3class}
\end{table}
\renewcommand{\arraystretch}{1}

\begin{table}[t]
\centering
\caption{Performance comparison between DiabNN-server and DiabNN-edge}
\begin{tabular}{ccc}
\hline
\hline
Performance matrices & DiabNN-server & DiabNN-edge\\
\hline
Accuracy                & 96.3\%            & 95.3\%    \\
FPR                     & 4.3\%             & 3.7\%    \\
FNR                     & 3.1\%             & 5.6\%    \\
F1                      & 96.5\%            & 95.4\%    \\
FLOPs                   & 858.2K            & 392.8K    \\
\#Parameters            & 429.1K            & 3.3K     \\
\hline
\hline
\label{tb:edge_server}
\end{tabular}
\end{table}

We also evaluate DiabNN-edge for the three-class classification task and present the confusion matrix 
in Table~\ref{tb:conf_DiabNN_edge_3class}. DiabNN-edge achieves an overall accuracy of 94.6\%. 
For the healthy case, it achieves an FPR of 4.5\%. It achieves FNRs of 6.5\% and 5.7\% for the 
type-1 and type-2 diabetic instances, respectively. DiabNN-server 
achieves an F1 score of 94.4\% for the three-class classification task.

DiabNN-edge delivers extreme model compactness. For binary classification, the final DiabNN-edge 
model contains a sparsity level of 96.3\%, yielding a model with only 3.3K 
parameters. For the three-class classification task, the final DiabNN-edge model 
contains a sparsity level of 95.9\%, yielding a model with only 3.7K 
parameters. This greatly assists with inference on the edge that 
typically suffers from very limited resource budgets.

\subsection{Results analysis}
\begin{figure}[!t]
\begin{center}
\subfigure[Change of accuracy and F1 score against model complexity.]{
\label{fig:acc_flops}
\includegraphics[width=9cm]{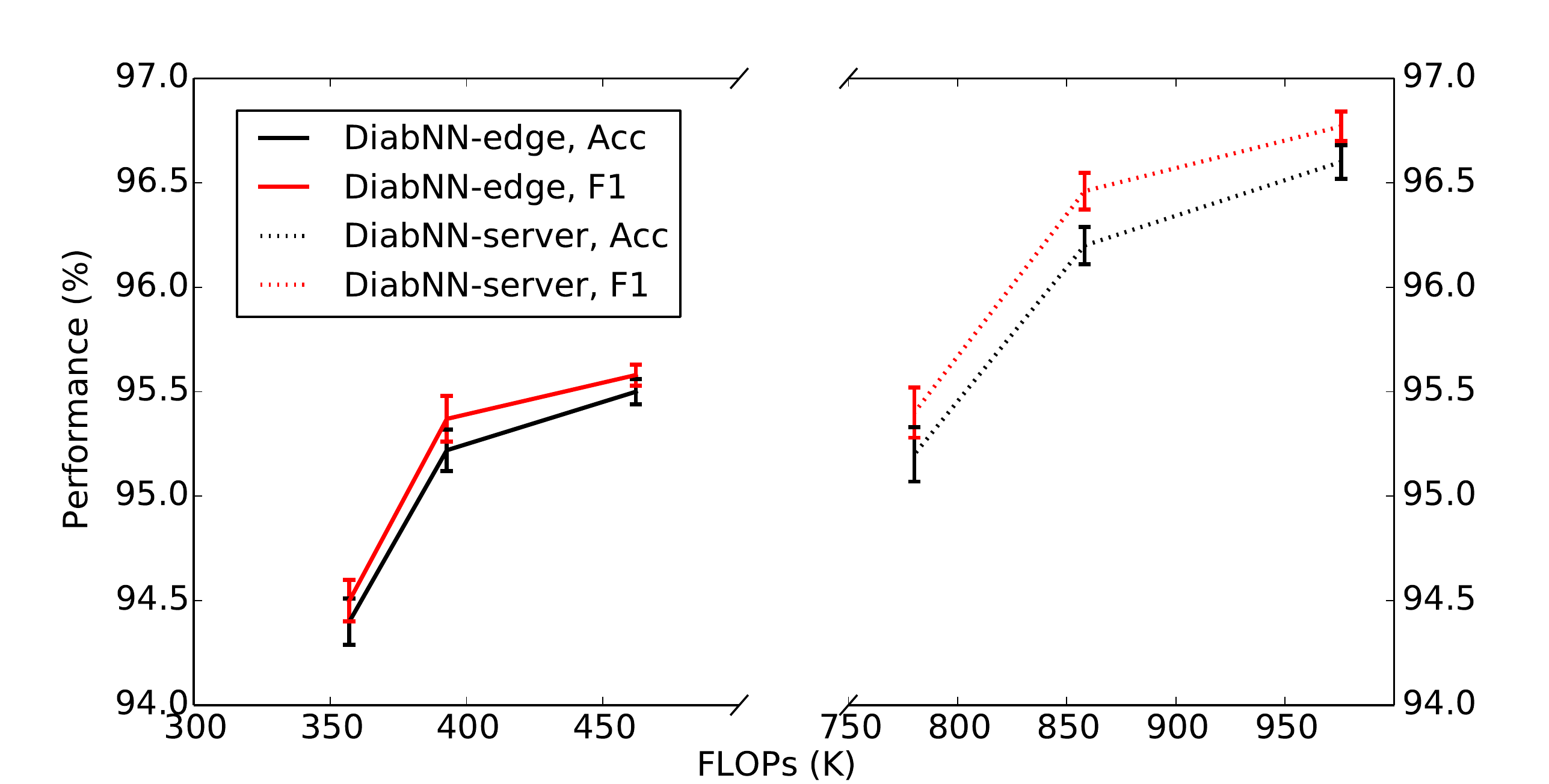}
}
\subfigure[Change of FPR and FNR against model complexity.]{
\label{fig:rl_nas_flow}
\includegraphics[width=9cm]{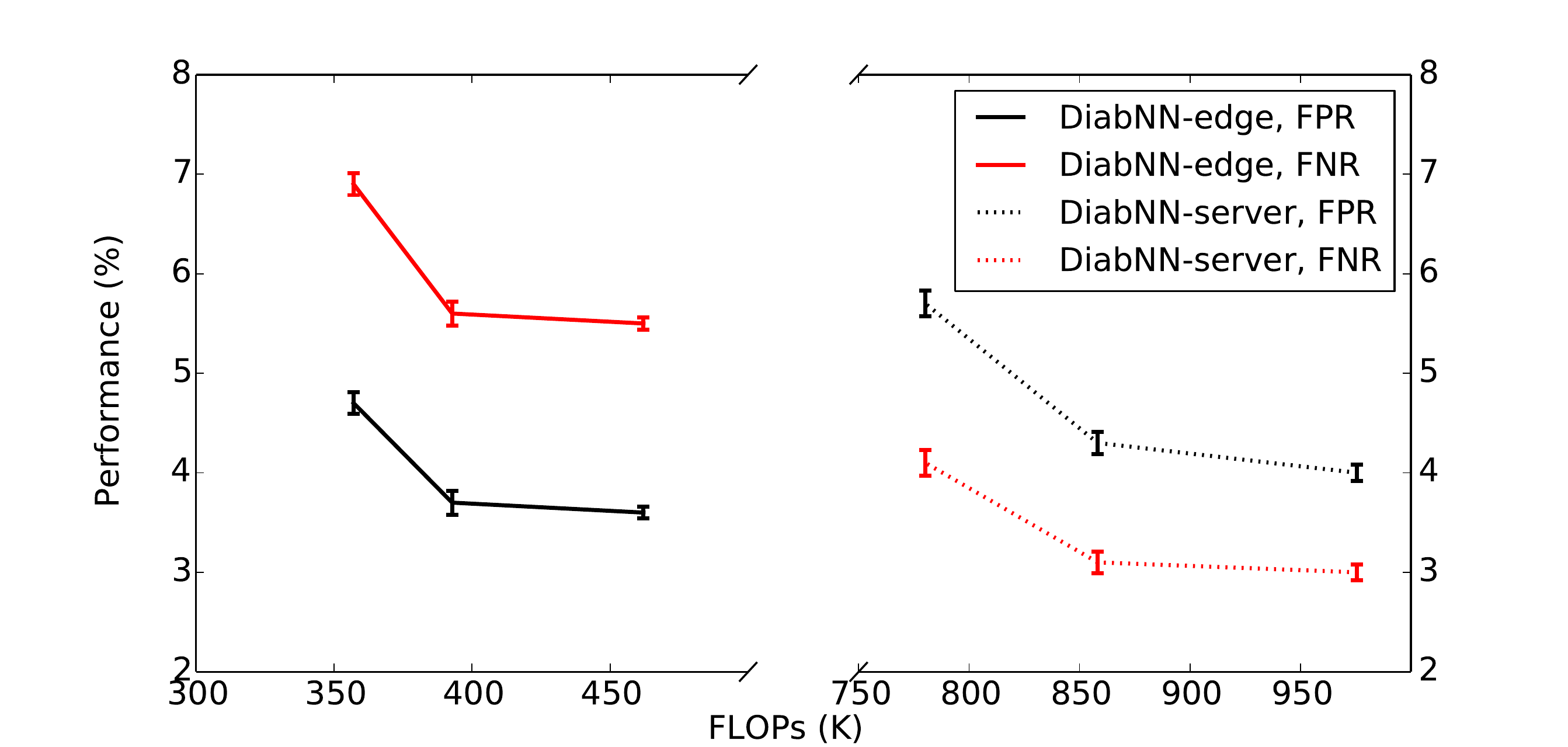}
}
\caption{Performance tradeoffs against model complexity based on five runs per data point with different random seeds. Error bars denote standard deviations.}
\label{fig:tradeoff}
\end{center}
\vspace{-14pt}
\end{figure}

As mentioned earlier, DiabNN-edge and DiabNN-server offer several performance tradeoffs over 
diagnostic accuracy, storage cost, and run-time efficiency. This provides flexible design choices 
that can accommodate varying design objectives related to model deployment. To illustrate their 
differences, we compare these two models for the binary classification task in 
Table~\ref{tb:edge_server}. We observe that DiabNN-server achieves a higher 
accuracy, a higher F1 score, and a 
lower FNR. DiabNN-edge, on the other hand, caters to edge-side inference by enabling:
\begin{itemize}
\item \textbf{A smaller model size}: The edge model contains $130\times$ fewer parameters, leading 
to a substantial memory reduction.
\item \textbf{Less computation}: It requires 2.2$\times$ fewer FLOPs per inference, enabling a more 
efficient, hence more frequent, monitoring capability on the edge.
\item \textbf{A lower FPR}: It reduces the FPR by 0.6$\%$. This enables fewer false alarms and 
hence an improved usability for the system in a daily usage scenario.
\end{itemize}
We also analyze the performance tradeoffs under changing model complexity 
and present the results in Fig.~\ref{fig:tradeoff}. It can be observed
that an increase in computational complexity can lead to performance
improvements, in general. However, such benefits gradually degrade as the computation complexity continues to increase. 

\begin{table*}[t]
\centering
\caption{Inference model comparison between DiabNNs and conventional methods}
\begin{tabular}{lrrrrr}
\hline
\hline
Model                   & Accuracy             & \#Parameters        & Feature extraction & Classification & Total FLOPs\\
& & & FLOPs & FLOPs & \\
\hline
SVM-linear              & 86.3\%        & 699K          & 0.49M & 1.39M & 1.88M \\
SVM-RBF                 & 92.1\%        & 822K          & 0.49M & 1.65M & 2.14M \\
k-NN                    & 93.5\%        & 1.5M          & 0.49M & 3.0M &  3.49M\\
Random forest           & 92.6\%         & 8K            & 0.49M & 4K$^+$ & 0.49M$^*$ \\
Linear ridge            & 81.3\%       & 0.8K          & 0.49M & 1.6K & 0.49M \\
\hline
DiabNN-server   & 96.3\% & 429K & - & 0.86M & 0.86M \\
DiabNN-edge     & 95.3\% & 3.3K & - & 0.39M & 0.39M \\
\hline
\hline
\multicolumn{6}{l}{\scriptsize{$+$: Number of comparison operations.}}\\
\multicolumn{6}{l}{\scriptsize{$*$: Calculation excluding the comparison operation cost.}}\\
\label{tb:energy_compare}
\end{tabular}
\end{table*}

\begin{table*}[t]
\centering
\caption{Performance comparison between DiabDeep and related work}
\begin{tabular}{lcccc}
\hline
\hline
 & Data sources & Model & Accuracy \\
\hline
Swapna et al.\cite{swapna2018automated}         & ECG sensor          &  Conv-LSTM         & 95.1\% \\
Swapna et al.\cite{swapna2018automated}        & ECG sensor                    &  CNN          & 93.6\% \\
Ballinger et al.~\cite{deepheart}          & Watch + demographics          &  LSTM         & 84.5\% \\
Yin et al.~\cite{hdss}   & Watch + demographics          &  Ensemble     & 77.6\% \\
\hline
This work (DiabNN-server)      & Watch + phone + demographics  &  Stacked SC layers          & 96.3\% \\
This work (DiabNN-edge)         & Watch + phone + demographics  &  H-LSTM SR layer        & 95.3\% \\
\hline
\hline
\label{tb:lit}
\end{tabular}
\end{table*}

We next compare DiabNNs with widely-used learning methods, including SVMs with linear and RBF 
kernels, k-NN, random forest, and linear ridge classifiers. For all the methods, we use the same 
train/validation/test split and the same binary classification task for a fair comparison. In line 
with the studies in~\cite{db} and~\cite{gupta2015performance}, we extract the signal mean, variance, 
Fourier transform coefficients, and the third-order Daubechies wavelet transform approximation and 
detail coefficients on Daubechies D2, D4, D8, and D24 filters from each monitoring window, resulting 
in a feature vector of length 304 per data instance. We train all our non-NN baselines using the 
Python-based Scikit learn libraries~\cite{scikit}. We compare the performance of all the inference 
models in Table~\ref{tb:energy_compare}. In addition to classification accuracy, we also compute the 
necessary FLOPs per inference involved in both feature extraction and classification stages. We
can see that DiabNN-server achieves the highest accuracy among all the models. With a higher accuracy 
than all the non-NN baselines, DiabNN-edge achieves the smallest model size (up to 454.5$\times$ 
reduction) and least FLOPs per inference (up to 8.9$\times$ reduction). Note that the feature 
extraction stage accounts for 491K FLOPs even before the classification stage starts executing. This 
is already 1.3$\times$ the total inference cost of DiabNN-edge. 

Finally, we compare DiabDeep with relevant work from the literature in Table~\ref{tb:lit}. We also 
focus on the same binary classification task that is the focus of these studies. DiabDeep achieves 
the highest accuracy relative to the baselines due to its two major advantages. First, it relies on 
a more comprehensive set of WMSs. This captures a wider spectrum of user signals in the daily context 
for diagnostic decisions. Moreover, it utilizes a grow-and-prune training paradigm that learns both 
the connections and weights in DiabNNs. This enables a more effective SGD in both the model 
architecture space and parameter space. 

\section{Discussions \& Future work}
\label{sec:discussions}
In this section, we discuss the inspirations we took from human brains
to train DiabNNs as well as the future directions enabled by DiabDeep.

Our brains continually remold the synaptic connections as we acquire new 
knowledge. These changes happen every second throughout our lifetimes. It 
has even been shown that most knowledge acquisition and information learning 
processes in our brains result from such a synaptic rewiring, also referred 
to as neuroplasticity~\cite{dilemma}. This is very different from 
most current NNs that have a fixed architecture. To mimic the learning 
mechanism of human brains, we utilize gradient-based growth and 
magnitude-based pruning to train accurate, yet very compact, DiabNNs for 
DiabDeep. The grow-and-prune synthesis paradigm allows DiabNNs to easily 
adjust their synaptic connections to the diabetes diagnosis task.

DiabDeep opens up the potential for future WMS-based disease diagnosis 
studies, given that more than 69,000 diseases exist~\cite{hdss}. We hope 
that this work will encourage clinics/hospitals/researchers to start 
collecting WMS data from individuals across more challenging diagnostic tasks, 
e.g., for long-term cancer prediction. Bypassing the feature extraction stage 
with efficient NNs enables easy scalability of the proposed approach across 
other disease domains. The grow-and-prune synthesis paradigm may even support 
continuous disease trend forecasting capability, given its continuous 
learning capability~\cite{incremental}. As more data become available and 
analyzed with the proposed methodology, its effectiveness as a scalable 
approach for future pervasive diagnosis and medication level 
determination will continue to improve.

\section{Conclusions}
\label{sec:conclusion}

In this work, we proposed a framework called DiabDeep that combines off-the-shelf WMSs with efficient 
DiabNNs for continuous and pervasive diabetes diagnosis on both the server and the edge. On the 
resource-rich server, we deployed stacked SC layers to focus on high accuracy. On the 
resource-scarce edge, we used an H-LSTM based SR layer to reduce computation and storage costs 
with only a minor accuracy loss. We trained DiabNNs by leveraging gradient-based growth and 
magnitude-based pruning algorithms. This enables DiabNNs to learn both weights and connections during 
training. We evaluated DiabDeep based on data collected from 52 participants. Our system achieves 
a 96.3\% (95.3\%) accuracy in classifying diabetics against healthy individuals on the server (edge), 
and a 95.7\% (94.6\%) accuracy in distinguishing among type-1 diabetic, type-2 diabetic, and healthy 
individuals. Against conventional baselines, such as SVMs with linear and RBF kernels, k-NN, random 
forest, and linear ridge classifiers, DiabNN-edge reduces model size (FLOPs) by up to 
454.5$\times$ (8.9$\times$) while improving accuracy. Thus, we have demonstrated that DiabDeep 
can be employed in a pervasive fashion, while offering high efficiency and accuracy.\\

\ifCLASSOPTIONcompsoc
  \vspace{-0.45cm}
  \section*{Acknowledgments}

The authors would like to thank Premal Kamdar, Abdullah Guler, Shrenik Shah, Aumify Health, and 
DiabetesSisters for assistance with data collection.

\ifCLASSOPTIONcaptionsoff
  \newpage
\fi

\bibliographystyle{IEEEtran} 
\bibliography{bibib} 

\vfill
\clearpage
\end{document}